# Small Language Models Also Work With Small Vocabularies: Probing the Linguistic Abilities of Grapheme- and Phoneme-Based Baby Llamas

**Bastian Bunzeck, Daniel Duran, Leonie Schade** and **Sina Zarrieß**
CRC 1646 – Linguistic Creativity in Communication
Department of Linguistics
Bielefeld University, Germany
`{firstname.lastname}`@uni-bielefeld.de

## Abstract

Recent work investigates whether LMs learn human-like linguistic generalizations and representations from developmentally plausible amounts of data. Yet, the basic linguistic units processed in these LMs are determined by subword-based tokenization, which limits their validity as models of learning at and below the word level. In this paper, we explore the potential of tokenization-free, phoneme- and grapheme-based language models. We demonstrate that small models based on the Llama architecture can achieve strong linguistic performance on standard syntactic and novel lexical/phonetic benchmarks when trained with character-level vocabularies. We further show that phoneme-based models almost match grapheme-based models in standard tasks and novel evaluations. Our findings suggest a promising direction for creating more linguistically plausible language models that are better suited for computational studies of language acquisition and processing.

## 1 Introduction

While very *large* language models possess human-like linguistic skills in many domains, questions of developmental plausibility have lead to a recent surge in interest for *small* language models (see, *inter alia*, Huebner et al., 2021; Warstadt et al., 2023; Choshen et al., 2024; Wilcox et al., 2024). Trained on more plausible amounts of data, these models are used to understand how different levels of linguistic knowledge are represented and learned. However, existing efforts mostly mostly test syntactic and syntacto-semantic capabilities (e.g. BLiMP, Warstadt et al., 2020). Representations and rules at lower levels of linguistic processing (phonology/morphology) cannot be meaningfully studied with such models, because common LMs adopt the same subword tokenization algorithms. Therefore, even developmentally oriented, so-called BabyLMs only learn representations for subwords, which are somewhat arbitrary (Uzan et al., 2024), do not (systematically) correspond to plausible linguistic units or segments like phonemes, syllables or morphemes (Beinborn and Pinter, 2023), and pose challenges for (psycho)linguistic modeling and investigations employing LMs (Giulianelli et al., 2024).

In this paper, we train and evaluate small Llama models (Touvron et al., 2023) on input that is not pre-segmented into words. Instead, we treat the individual characters in our training data as tokens. Therefore, our models do not receive any prior information on what "meaningful" units in the input are. We investigate whether these small models, trained with drastically smaller and linguistically more plausible vocabularies, still achieve comparable performance on evaluations across the phonetic, the lexical and the syntactic level. Additionally, we compare models trained on graphemes and models trained on phonemes[1], questioning the common assumption that grapheme-based models are as *tabula rasa* (Hahn and Baroni, 2019) as LMs can get.

We find that our character-based LMs perform as well on standard evaluation measures as comparable subword-based models trained on the same data. We also show that our models are able to learn representations of lexical and phonological units surprisingly well, outperforming subword models on a lexical decision task. Further, we find that phoneme-based models generally show worse performance on syntactic and lexical evaluations, but do perform on par with grapheme-based models in rhyme prediction and even show advantages in our speaker age prediction task. This suggests that training data for grapheme-based models comes with more inherent structural "biases" than commonly assumed, e.g. in the form of punctuation. However, these biases are not conducive to every linguistic task, as the speaker age prediction task shows.

---

[1]The terms (i) "grapheme" and (ii) "phoneme" are used loosely in this paper to refer to (i) individual characters in a language's alphabet and (ii) types of speech sounds, in accordance with the G2P literature (cf. Moore and Skidmore, 2019; Ashby et al., 2021).

## 2 Related work

Several approaches are being currently explored to train language models (LMs) without relying on subword tokenization, aiming to achieve more naturalistic and linguistically plausible representations. Notably, these include: (i) end-to-end training on raw speech signals (Chrupała, 2022; Nguyen et al., 2023b; Vong et al., 2024), and (ii) grapheme-level LMs where tokens correspond to characters of the respective language script. (i) is fully naturalistic, but limited by data availability and a lack of evaluation protocols (Nikolaus et al., 2022), while (ii) foregoes the complexities of audio processing, but relies on orthography which is rarely a good approximation of actual pronunciation. In this paper, we explore a third approach that bridges these two paradigms by training on sequences of phoneme symbols, derived from transcriptions or (in our case) grapheme-to-phoneme (G2P) conversion.

**Character-level LMs:** Subword-less or character-level LMs have so far mostly been used in pre-training for downstream tasks: CANINE (Clark et al., 2022) employs a character-level encoder architecture while Xue et al. (2022) adapt a T5 encoder-decoder architecture to work on the Byte-level. Both models show that pre-training on characters can provide useful inductive biases and lead to models that are more robust to noise than subword models. Similarly, the Charformer architecture (Tay et al., 2022) performs on par with comparable subword-based models despite reduced computational cost. These findings are also corroborated by Cao (2023), who provide an overview of extensive experiments and ablations with character-based encoders. The most similar approach to ours (i.e. most linguistically oriented) is the one by Hahn and Baroni (2019), who train character-level RNNs as *tabula rasa* learners and find that they learn morphological, syntactic and lexical properties of language from unsegmented input. For current, state-of-the-art Transformer LMs (e.g. Llama, GPT, etc.), only few results on this specific training regimen exist (Goriely et al., 2024; Bunzeck et al., 2024).

**Phoneme-based LMs:** Research on phoneme-based LMs is also fairly limited. Phoneme-BERT (Sundararaman et al., 2021), trained on joint grapheme and phoneme representations, has been shown to provide useful inductive biases for further fine-tuning on downstream tasks from speech technology. Similar results are reported for Mixed-Phoneme BERT (Zhang et al., 2022) and XPhoneBERT (Nguyen et al., 2023a) on text-to-speech tasks, and for BORT (Gale et al., 2023) on clinical tasks that rely on phonetic and semantic information. Like the character-based models reviewed above, these phoneme-based models are united by being encoder models that are only used as pre-training for downstream tasks and not for linguistic investigations. Futhermore, none of these models operate on a (naïve) character-level. From a linguistic viewpoint, Nguyen et al. (2022) showed that grapheme-based LSTMs are superior to phoneme-based LSTMs on lexical, syntactic and semantic tasks, both Goriely et al. (2024) and Bunzeck et al. (2024) showed the same for autoregressive Transformer models (GPT-2 and Llama, respectively) in the context of the 2024 BabyLM challenge (Choshen et al., 2024).

**Phonetic/phonological benchmarks:** Similarly to the lack of phoneme-based LMs, benchmarks on phonetics/phonology are rare to non-existent. Suvarna et al. (2024) propose three different tasks for their phonology benchmark – grapheme-to-phoneme conversion (G2P), syllable counting and rhyme generation. By relying solely on prompting strategies to explicitly test the generative capabilities of LLMs, their results fail to provide insight into the underlying representations these models have learned (Hu and Levy, 2023). Lavechin et al. (2023) provide a speech-synthesized version of a lexical decision task and a syntactic minimal pair task. In absence of more targeted phonetic evaluations, we first employ the BabyLM evaluation protocol. Furthermore, we expand it with a lexical decision task (similar to Lavechin et al., 2023, but with comparable data on the grapheme and phoneme level), and add two more phonological/phonetic tasks – rhyme prediction and age prediction from sequence embeddings.

## 3 Methodology

### 3.1 Data and pre-training

All our models[2] are trained on the same 100M-token BabyLM 2024 data set (see Choshen et al.,

---
[2] Details such as model hyperparameters and Huggingface links can be found in Appendix A.1.

2024 for composition and included corpora), which we preprocess in different ways for each model. For the phoneme models, we convert the cleaned data from graphemes to phonemes with $G_i2P_i$ (Pine et al., 2022)³. As $G_i2P_i$ fails to correctly convert contractions, we add transcription rules for such forms manually. For both data sets (grapheme and phoneme), we train one model on the raw data set and one model on the data set without white spaces. Through this, we aim to remove another structural property of written language (visible word boundaries), which does not correspond well to any property of spoken language. All models are comparatively "small" (approx. 15M parameters).

We implement character-level language modeling by modifying the tokenizers to only include the set of unique characters in the respective pre-training corpus. For the grapheme-based models, this adds up to a vocabulary size of approx. 360. For the phoneme models, the vocabulary size is approx. 260 (including "noise" like emojis and other irrelevant characters). As an ablation, we also evaluate the BabyLM 2024 baseline model BabyLlama (Timiryasov and Tastet, 2023), which uses regular subword tokenization (vocab. size of 16.000) and has almost four times more parameters (58M) than our models.

### 3.2 Evaluation

Table 1 gives a short overview of our evaluation approaches and the data used. We share our newly created data on Huggingface (see Appendix A.3 for details).

| Example (graphemic) | Example (phonetic) |
|---|---|
| **BLiMP** (Minimal pairs) | |
| 👍 Aaron breaks the glass. | 👍 ɛɹʌn bɹeɪks ðʌ glæs |
| 👎 Aaron appeared the glass. | 👎 ɛɹʌn ʌpɪɹd ðʌ glæs |
| **Lexical decision task** (Minimal pairs) | |
| 👍 drunk. | 👍 dɹʌŋk |
| 👎 blunk. | 👎 fɹʌŋk |
| **Rhyme prediction** (Probing) | |
| ✔ The sky was clear, but full of cheer. | ✔ ðʌ skaɪ wɑz klɪɹ bʌt fʊl ʌv tʃɪɹ |
| ✘ The door opened with a creak. | ✘ ðʌ dɔɹ oʊpʌnd wɪð ʌ kɹik |
| **Age prediction** (Probing) | |
| 👶 rock , rock , rock . | 👶 waːwaːwaː |
| 🧒 hold my juice Mommy . | 🧒 hod maɪ dʒus mami |
| 👦 open the door . | 👦 opən ðə dɔr |

Table 1: Examples of all evaluation paradigms

**BLiMP:** For general evaluation we use BLiMP (Warstadt et al., 2020) and supplementary BLiMP

---
³See Appendix A.2 for an evaluation of transcription quality, and Appendix A.3 for links to our data.

data from the BabyLM challenge. BLiMP is a widely adopted minimal pair dataset for English, which features over 70 linguistic paradigms (sets of specific instantiations of linguistic phenomena). The phenomena are mostly taken from syntax, but also include morphosyntax, semantics, and some dialogue phenomena. We compute perplexities (Jelinek et al., 1977) for grammatical and ungrammatical sentences and report preference scores for the grammatical utterances. For our phoneme-based models, we create Phoneme-BLiMP using the same G2P procedure as for our training data and evaluate the models on this data set instead.

**Lexical decision task:** Since BLiMP hardly includes any tasks at/below the word level, we use a lexical decision – a common testing paradigm in psycholinguistics, see (Perea et al., 2005; Yap et al., 2015) – to assess lexical capabilities. Inspired by Le Godais et al. (2017), we use `wuggy` (Keuleers and Brysbaert, 2010) to generate nonce words from 1,000 English high-frequency words. `wuggy` generates nonce words that contain plausible character/phoneme sequences while trying to preserve bigram frequencies inside the words as accurately as possible. That being the case, the concrete stimuli should not constitute a confounding factor. Grapheme and phoneme nonce words were generated separately to avoid creating orthographic nonce words that on a sound level correspond to existing words. Here, we also compute perplexity scores for all stimuli and compare whether models prefer the existing word or the nonce word.

**Rhyme prediction:** As rhyming is an important property of phonology (e.g. in acquisition, see Goswami, 2001), we employ a probing approach (Belinkov, 2022) to assess if LMs do encode rhymes. We create a small (200 sentences) data set with in-context learning (Dong et al., 2024) through ChatGPT-4o. From this dataset, we create sentence embeddings with our models. From these embeddings, we train a linear regression classifier and report accuracy scores after ten-fold cross validation.

**Age prediction:** Finally, we use the same probing approach for the prediction of child age from utterances. This is an interesting task because the phonological properties of child language and childrens' usable phone inventories change drastically across the first years of development (Saxton, 2017). We take an age-balanced sample of 1.000 child ut-terances from the Providence (Demuth et al., 2006) and ComptonPater (Compton and Streeter, 1977; Pater, 1997) corpora, which contain parallel, hand-crafted phonetic transcriptions for the orthographic data. As the following age groups mark important developmental milestones, we use the utterance embeddings to categorize speaker age as below 1 (production of first utterances, Schneider et al., 2015), 1–3 (rapid vocabulary spurt), and over 3 (stabilizing vocabulary, cf. Frank et al., 2017, 2021). We report accuracy scores after ten-fold cross validation.

| Evaluation | Grapheme model | Grapheme model, no whitesp. | Phoneme model | Phoneme model, no whitesp. | BabyLlama |
|---|---|---|---|---|---|
| BLiMP | 71.69% | 68.88% | 66.90% | 64.88% | 73.10% |
| BLiMP supplement | 52.30% | 56.28% | 55.42% | 54.13% | 60.60% |
| Lexical decision task | 99.00% | 99.10% | 68.20% | 63.80% | 69.00% |
| Rhyme prediction | 88.50% | 91.50% | 85.00% | 78.49% | 92.50% |
| Age prediction | 60.50% | 58.90% | 61.10% | 57.80% | 60.90% |

Table 2: Evaluation results: for BLiMP and the lexical decision task, the scores correspond to the percentage of correct choices in a minimal pair setting; for rhyme and age prediction the scores report classification accuracy.

## 4 Results

The results of all evaluations are shown in Table 2.

**BLiMP:** The grapheme model achieves a BLiMP score of over 70% and performs similarly to the subword-based BabyLlama. The other models perform 3–7% worse, with a decrease for phoneme models and for non-whitespace models. Interestingly, the phoneme models drastically outperform the grapheme models on a few phenomena, e.g. sentential_subject_island_filtered (see full BLiMP scores in Appendix B). On the supplementary phenomena, the subword-based model achieves the best score, whereas our models only perform slightly above chance.

**Lexical decision task:** Grapheme models achieve almost perfect performance on the lexical decision task – in fact, the model trained without word-separating white space beats its counterpart by a tiny margin. While the phoneme models achieve worse scores (63–68%), they still perform above chance and in the same range as the subword-based BabyLlama, which fares much worse than its character-based equivalent (despite featuring a much higher number of parameters).

**Rhyme prediction:** For the rhyme prediction task, both the grapheme and phoneme models perform well above the chance baseline. Again, the grapheme models surpass the phoneme models moderately and achieve scores in the same range as BabyLlama. While the deletion of whitespace has a detrimental effect for the phoneme model, it improves the score for the grapheme model. This might indicate that our grapheme model trained without white space characters is pushed towards learning more precise representations at the lexical/phonological level.

**Age prediction:** The age prediction task shows roughly equivalent results between grapheme/phoneme models and BabyLlama. All models achieve scores around approx. 60% and therefore perform well above the chance baseline (here: 33%). In fact, this is the only task where the phoneme model beats its grapheme counterpart by a tiny margin (0.6%) and features the overall best performance of all evaluated models.

In sum, the following trends are identifiable: (i) The most striking result is that the subword-based BabyLlama does not generally perform better on linguistic benchmarks than character-based models. While it performs best on the syntactic evaluation, it struggles with lexical decision. On the phonetic prediction tasks, it performs similarly to character-based models. (ii) Grapheme models are generally superior to phoneme models, but the differences are less pronounced or barely noticeable for the more phonetically inclined rhyme and age prediction tasks. (iii) The deletion of white spaces has a negative effect on syntactic evaluations, but moderately improves the grapheme models on the BLiMP supplement and the rhyme prediction task. For the phoneme models, the removal of whitespace has a consistently detrimental effect.

## 5 Discussion

Several conclusions follow from our current, preliminary results: (i) Our character-level LMs perform as well as (larger) subword-based LMs on syntactic tests and embedding-based prediction tasks. On a lexical decision task, they even surpass them. While Clark et al. (2022) report how naïve character-based CANINE models perform worse on linguistic tasks, our results put these findings into question and align more with recent findings that character representations do not harm performance tremendously (Xue et al., 2022; Kodner et al., 2023; Goriely et al., 2024; Bunzeck et al., 2024). When combined with state-of-the-art architectures like Llama, character-based models work extremely well and excel in tasks that sub-word models are not suited to (like lexical decision). (ii) Phoneme LMs are also able to capture linguistic phenomena, but they generally perform worse than grapheme models. However, they surpass grapheme models on certain, specific tasks. This aligns with Kodner et al. (2023), who argue the phoneme models should be the default for cognitive investigations with LMs. (iii) Models trained without white space show moderate improvements at lexical/phonological tasks, which might indicate that grapheme models trained without white space characters actually develop more precise and granular latent representations at the lexical and phonological levels. By excluding explicit word boundaries, these models are likely forced to infer word-level and subword-level structures from the data itself, potentially leading to a deeper encoding of phonological and lexical patterns.

The systematic superiority of grapheme- over phoneme-based models calls the commonly assumed *tabula rasa*-ness of grapheme-based models (Hahn and Baroni, 2019) into question. Explaining these effects requires further research and we believe the following directions to be worth exploring: (i) grapheme-based models may pick up all kinds of inductive biases introduced by orthography (for example, punctuation marks transport information about word and clause boundaries, whereas the grapheme <ght> signals a syllable boundary, which is not the case for the corresponding /t/-sound in phonetic transcriptions), (ii) phoneme-based models may suffer from errors introduced through too rigid G2P. G2P tools are commonly based on pronunciation dictionaries, which include *broad* transcriptions (e.g. /ˈtu/ for *'two'*). They transcribe to "canonical forms" of pronunciation, while in reality pronunciation depends on the phonological and situative context, individual/social factors, etc., and is much more varied. Phoneme models could benefit from more fine-grained representations, particularly for phonological tasks requiring subtle distinctions in pronunciation. However, training on phonetically plausible data requires manually checked transcriptions, which are rare and often limited in size (e.g., the PhonBank section of CHILDES (MacWhinney, 2000)). Currently, G2P-converted data lacks the advantages of grapheme data and the variability of real-world phonetic data, making it less effective for learning.

# 6 Conclusion

This paper has demonstrated (as a *proof-of-concept*) that character-based grapheme and phoneme models can capture linguistic structures effectively and offer valuable insights into linguistic learning that are difficult to derive from subword-based models. Notably, these models show advantages on specific tasks, such as lexical decision, where their performance quite drastically exceeds that of their subword-based counterparts. This finding underscores that subword tokenization – which is the de-facto standard in current language modeling practices, but is not grounded in meaningful linguistic assumptions – obfuscates basic dimensions of language learning in LMs that happen at and below the word level.

The observed weaker performance of phoneme models on other tasks remains an intriguing issue. Future work should investigate whether and how these models develop representations of higher linguistic units, such as syllables or morphemes, and how their latent vector spaces differ from those of subword-based models. Moreover, our results suggest that character-based tokenization may compel models to encode more precise lexical or phonological patterns, a hypothesis that warrants further exploration with directed experiments. Finally, it is important to emphasize that the primary aim of this study is not to maximize performance on real-world applications or downstream tasks, but to systematically examine how representation and tokenization choices influence linguistic generalization in small-scale LMs. As such, this work aligns with the BabyLM tradition of exploring the emergence of linguistic knowledge in constrained settings, providing a foundation for future research into the cognitive and representational properties of (possibly larger) LMs.

# Limitations

Our study is constrained by three factors: (i) The choice of data and G2P tool, which enforces a specific type of transcription, could also influence how the linguistic system is formed in our LMs. More information and (naturalistic) variation could possibly lead to different (dis)advantages across bench-marks. (ii) Our choice of architecture. We use an autoregressive decoder, as these models are currently the state-of-the-art for language modeling, but it remains open as to whether encoder-only or encoder-decoder models learn similar representations. (iii) The availability of benchmarks. As subword-based models do not feature representations for phonetic/phonological units, probing and evaluation approaches have so far not focused on these linguistic levels. More diverse benchmarking options are needed to fully evaluate shortcomings and advantages of character-level and/or phoneme models. All of these factors deserve further research and consideration.

## Ethical considerations

We acknowledge that the "standardization" of pronunciation that our G2P tool enforces is rather exclusionary towards variation. While orthographic conventions convey comparably little information about the variety of (standard) English, as in *categorise* vs *categorize*, pronunciation conveys social and individual information about the speaker's identity and linguistic background. The systematic G2P conversion of text data does not include such variability. As such, the approach implemented here is impacted by non-inclusivity, as it is biased towards a standard variety and ignores, e.g., sociolinguistic variation (Schubert et al., 2024). Corpora of transcribed spontaneous speech would provide a more diverse representation of a language, since they may include regional, social and individual variation (Schweitzer et al., 2015).

If our models were used in an applied setting, they definitely need more of such variation as input. Speech technology, e.g. voice assistants, produce (and recognize) almost only standard, i.e. non-dialectal/accented speech. Most people around the world use situation-specific non-standard varieties and speaking styles in everyday communication. Furthermore, most of the world's population is bilingual (Wei, 2005). Code-switching between languages inside one utterance is an integral part or their everyday communication. Our approach does not handle such kinds of variation (yet).

## Acknowledgements

This research has been funded by the Deutsche Forschungsgemeinschaft (DFG, German Research Foundation) – CRC-1646, project number 512393437, project A02.

## A Technical details

### A.1 Model details

All our models are available on the Huggingface hub:

https://huggingface.co/bbunzeck/grapheme-llama
https://huggingface.co/bbunzeck/grapheme-llama-no-whitespace
https://huggingface.co/bbunzeck/phoneme-llama
https://huggingface.co/bbunzeck/phoneme-llama-no-whitespace

They share equivalent model internals/hyperparameters: 8 hidden layers, 8 attention heads, an embedding size of 512 and a short context size of 64.

Psychological models of speech processing assume that working memory and attention are limited cognitive resources (e.g. Baddeley, 2003, 2017; Cowan, 2016). Often, a number of approximately seven items is assumed which can be held in working memory (Miller, 1956) — which is considerably smaller than our very short context size. We train all models for five epochs.

### A.2 G2P processing

Because Pine et al. (2022) report no G2P accuracy for English, we conduct a manual evaluation on three short texts, including the standard text "The Northwind and the Sun" (International Phonetic Association, 1999); and two short samples of movie subtitles for "PAW Patrol: The Mighty Movie" and "The SpongeBob Movie: Sponge Out of Water" [https://www.opensubtitles.org]. We find a WER of 5.8% (tokens=363, errors=21). $G_i2P_i$ features several shortcomings: as a rule-based system, it cannot handle *creative* words (e.g. *"Just keep weading Pwease Mr Piwate sir"*). Also it does not mark stress like some other G2P tools. For our current purposes, however, we deem the performance of $G_i2P_i$ as sufficient.

### A.3 Training and evaluation data

We share our newly created data sets on the Huggingface hub:

The converted BabyLM data sets can be found at https://huggingface.co/datasets/bbunzeck/phoneme-babylm-10M and https://huggingface.co/datasets/bbunzeck/phoneme-babylm-100M.

PhonemeBLiMP is available at https://huggingface.co/datasets/bbunzeck/phoneme-blimp.
The lexical decision data is available at https://huggingface.co/datasets/bbunzeck/wug-words.
The rhyme data can be found at https://huggingface.co/datasets/bbunzeck/rhyme-sentences.

The age prediction data was extracted from the CHILDES corpora available at https://phon.talkbank.org/access/Eng-NA/ComptonPater.html and https://phon.talkbank.org/access/Eng-NA/Providence.html.

# B  Full BLiMP scores

| Phenomenon | Graph. model | Graph. model, no whitesp. | Phon. model | Phon. model, no whitesp. |
|---|---|---|---|---|
| BLiMP | **71.69%** | 68.88% | 66.90% | 64.88% |
| BLiMP supplement | 52.30% | **56.28%** | 55.42% | 54.13% |
| adjunct_island_filtered | 73.17% | **76.72%** | 35.24% | 36.75% |
| anaphor_gender_agreement_filtered | 85.48% | 82.29% | **86.30%** | 69.10% |
| anaphor_number_agreement_filtered | **97.10%** | 88.51% | 95.17% | 87.00% |
| animate_subject_passive_filtered | 68.60% | **71.62%** | 68.83% | 62.91% |
| animate_subject_trans_filtered | **91.01%** | 90.57% | 82.23% | 77.79% |
| causative_filtered | **69.07%** | 68.09% | 66.01% | 64.55% |
| complex_NP_island_filtered | 43.38% | **47.28%** | 38.30% | 43.85% |
| coordinate_structure_constraint_complex_left_branch_filtered | **46.36%** | 37.75% | 36.31% | 30.68% |
| coordinate_structure_constraint_object_extraction_filtered | 62.38% | 65.12% | **65.86%** | 63.22% |
| determiner_noun_agreement_1_filtered | 97.31% | **97.74%** | 52.85% | 52.85% |
| determiner_noun_agreement_2_filtered | 96.99% | **97.10%** | 85.61% | 82.81% |
| determiner_noun_agreement_irregular_1_filtered | **83.85%** | 78.12% | 72.25% | 70.78% |
| determiner_noun_agreement_irregular_2_filtered | **90.00%** | 87.56% | 84.15% | 76.59% |
| determiner_noun_agreement_with_adj_2_filtered | **92.24%** | 90.75% | 79.81% | 76.94% |
| determiner_noun_agreement_with_adj_irregular_1_filtered | **82.45%** | 77.30% | 73.96% | 71.17% |
| determiner_noun_agreement_with_adj_irregular_2_filtered | **82.38%** | 78.93% | 72.26% | 69.88% |
| determiner_noun_agreement_with_adjective_1_filtered | **94.96%** | 91.00% | 51.77% | 51.55% |
| distractor_agreement_relational_noun_filtered | **86.29%** | 45.05% | 68.40% | 57.11% |
| distractor_agreement_relative_clause_filtered | **58.09%** | 43.17% | 50.98% | 57.41% |
| drop_argument_filtered | 75.76% | **75.98%** | 60.87% | 62.07% |
| ellipsis_n_bar_1_filtered | 51.50% | **56.36%** | 54.36% | 53.87% |
| ellipsis_n_bar_2_filtered | 58.09% | **63.29%** | 43.36% | 49.64% |
| existential_there_object_raising_filtered | **81.65%** | 72.66% | 79.80% | 68.10% |
| existential_there_quantifiers_1_filtered | **99.46%** | 97.42% | 96.77% | 93.76% |
| existential_there_quantifiers_2_filtered | 28.21% | 33.92% | 38.42% | **43.69%** |
| existential_there_subject_raising_filtered | 83.98% | 82.90% | **84.31%** | 80.84% |
| expletive_it_object_raising_filtered | 70.09% | **73.12%** | 72.46% | 70.22% |
| inchoative_filtered | **55.79%** | 52.28% | 44.91% | 46.67% |
| intransitive_filtered | **68.32%** | 67.17% | 46.31% | 50.58% |
| irregular_past_participle_adjectives_filtered | **94.80%** | 88.14% | 72.84% | 63.58% |
| irregular_past_participle_verbs_filtered | 81.53% | 81.10% | **85.14%** | 77.39% |
| irregular_plural_subject_verb_agreement_1_filtered | **83.33%** | 76.62% | 82.21% | 72.14% |
| irregular_plural_subject_verb_agreement_2_filtered | **89.46%** | 87.33% | 88.00% | 83.86% |
| left_branch_island_echo_question_filtered | 65.15% | 61.67% | 63.15% | **70.86%** |
| left_branch_island_simple_question_filtered | **60.15%** | 46.79% | 57.83% | 50.26% |
| matrix_question_npi_licensor_present_filtered | 15.82% | 12.38% | 17.98% | **31.75%** |
| npi_present_1_filtered | **50.39%** | 40.59% | 46.75% | 48.51% |
| npi_present_2_filtered | 49.89% | **50.33%** | 45.62% | 48.69% |
| only_npi_licensor_present_filtered | **98.07%** | 48.64% | 76.87% | 92.06% |
| only_npi_scope_filtered | 50.90% | 44.92% | 61.05% | **80.53%** |
| passive_1_filtered | 89.17% | **90.60%** | 87.74% | 86.79% |
| passive_2_filtered | 88.15% | **89.37%** | 83.61% | 81.28% |
| principle_A_c_command_filtered | 55.07% | **59.51%** | 51.48% | 59.41% |
| principle_A_case_1_filtered | **100.00%** | **100.00%** | **100.00%** | 99.89% |
| principle_A_case_2_filtered | 91.58% | **92.57%** | 88.20% | 78.80% |
| principle_A_domain_1_filtered | 96.39% | 98.25% | **100.00%** | **100.00%** |
| principle_A_domain_2_filtered | 53.55% | 50.71% | **63.61%** | 51.80% |
| principle_A_domain_3_filtered | 50.90% | 50.90% | **61.00%** | 55.58% |
| principle_A_reconstruction_filtered | 41.88% | 34.64% | **53.67%** | 47.67% |
| regular_plural_subject_verb_agreement_1_filtered | **93.48%** | 90.45% | 88.76% | 80.11% |
| regular_plural_subject_verb_agreement_2_filtered | **90.37%** | 85.19% | 82.65% | 77.67% |
| sentential_negation_npi_licensor_present_filtered | 96.19% | 96.74% | **99.35%** | 96.52% |
| sentential_negation_npi_scope_filtered | 21.70% | 23.08% | 33.30% | **40.76%** |
| sentential_subject_island_filtered | 40.89% | 39.33% | **58.17%** | 57.54% |
| superlative_quantifiers_1_filtered | 66.70% | 66.80% | **70.99%** | 54.14% |
| superlative_quantifiers_2_filtered | 76.37% | **83.77%** | 69.98% | 61.16% |
| tough_vs_raising_1_filtered | **36.50%** | 28.80% | 23.73% | 29.32% |
| tough_vs_raising_2_filtered | 81.41% | **82.93%** | 80.76% | 78.37% |
| transitive_filtered | **80.07%** | 74.77% | 70.85% | 66.94% |
| wh_island_filtered | 61.77% | **63.54%** | 61.04% | 38.75% |
| wh_questions_object_gap_filtered | 78.70% | 75.20% | **80.33%** | 76.37% |
| wh_questions_subject_gap_filtered | 92.32% | **92.54%** | 92.43% | 90.31% |
| wh_questions_subject_gap_long_distance_filtered | 91.60% | 93.35% | 93.58% | **94.87%** |
| wh_vs_that_no_gap_filtered | 95.82% | 95.93% | **96.17%** | 94.54% |
| wh_vs_that_no_gap_long_distance_filtered | 94.86% | **97.37%** | 96.57% | 94.74% |
| wh_vs_that_with_gap_filtered | **27.20%** | 26.01% | 5.55% | 7.07% |
| wh_vs_that_with_gap_long_distance_filtered | **7.03%** | 4.18% | 3.41% | 4.62% |
| supplement_hypernym | 51.19% | **51.90%** | 51.07% | 51.19% |
| supplement_qa_congruence_easy | 48.44% | 54.69% | 56.25% | **57.81%** |
| supplement_qa_congruence_tricky | 26.67% | **39.39%** | 25.45% | 25.45% |
| supplement_subject_aux_inversion | 78.54% | 77.22% | **86.11%** | 79.75% |
| supplement_turn_taking | 56.79% | **58.21%** | **58.21%** | 56.43% |